\documentclass{article} 
\usepackage[preprint]{nips21}
\usepackage{times}
\usepackage[utf8]{inputenc} 
\usepackage[T1]{fontenc}    
\usepackage{hyperref}       
\usepackage{url}            
\usepackage{booktabs}       
\usepackage{amsfonts}       
\usepackage{nicefrac}       
\usepackage{microtype}      
\usepackage{xcolor}         
\usepackage{natbib}
\usepackage{xspace}
\usepackage{dsfont}
\usepackage{amsmath,amssymb}
\usepackage[ruled,norelsize]{algorithm2e}
\usepackage{multirow}
\usepackage{booktabs}
\usepackage{graphicx}
\usepackage{subcaption}
\usepackage{wrapfig}
\usepackage{enumerate}
\usepackage{enumitem}
\usepackage{url}
\usepackage{setspace}
\usepackage{arydshln}
\usepackage{hyperref}

\usepackage{amsmath,amsfonts,bm}

\newcommand{\eq}[1]{Eq.(\ref{#1})}
\newcommand{\anB}[1]{\langle #1 \rangle}

\def\va{{\bm{a}}}

\def\vc{{\bm{c}}}

\def\vh{{\bm{h}}}

\def\vm{{\bm{m}}}

\def\vr{{\bm{r}}}

\def\vx{{\bm{x}}}

\def\mM{{\bm{M}}}

\DeclareMathAlphabet{\mathsfit}{\encodingdefault}{\sfdefault}{m}{sl}
\SetMathAlphabet{\mathsfit}{bold}{\encodingdefault}{\sfdefault}{bx}{n}

\def\cL{{\mathcal{L}}}

\def\bR{{\mathbb{R}}}

\newif\ifcomments
\commentstrue

\newcommand{\ttt}[1]{\texttt{#1}}

\newcommand{\llm}{f_\text{llm}}
\newcommand{\mem}{\ttt{<mem>}}
\newcommand{\ret}{\ttt{<ret>}}

\SetKwInput{KwInit}{Init}

\title{The Compressor-Retriever Architecture for Language Model OS}

\author{Yuan Yang\textsuperscript{1,*}, Siheng Xiong\textsuperscript{1}, Ehsan Shareghi\textsuperscript{2} \& Faramarz Fekri\textsuperscript{1}\\
\textsuperscript{1}Georgia Institute of Technology, 
\textsuperscript{2}Monash University \\
\textsuperscript{*}\ttt{mblackout@hotmail.com} \\
\ttt{\{sxiong45@,faramarz.fekri@ece.\}gatech.edu} \\
\ttt{ehsan.shareghi@monash.edu}
}

\begin{document}

\maketitle

\begin{abstract}

Recent advancements in large language models (LLMs) have significantly enhanced their capacity to aggregate and process information across multiple modalities, enabling them to perform a wide range of tasks such as multimodal data querying, tool usage, web interactions, and handling long documents.
These capabilities pave the way for transforming LLMs from mere chatbots into general-purpose agents capable of interacting with the real world.
This paper explores the concept of using a language model as the core component of an operating system (OS), effectively acting as a CPU that processes data stored in a context window, which functions as RAM.
A key challenge in realizing such an LM OS is managing the life-long context and ensuring statefulness across sessions, a feature limited by the current session-based interaction paradigm due to context window size limit.
To address this, we introduce compressor-retriever, a model-agnostic architecture designed for life-long context management.
Unlike other long-context solutions such as retrieval-augmented generation, our approach exclusively uses the base model's forward function to compress and retrieve context, ensuring end-to-end differentiability.
Preliminary experiments demonstrate the effectiveness of this architecture in in-context learning tasks, marking a step towards the development of a fully stateful LLM OS.
Project repo available at:
\href{https://github.com/gblackout/LM-OS}{https://github.com/gblackout/LM-OS}

\end{abstract}

\section{Introduction}


LLMs demonstrate a strong capability as a central model that aggregates and processes information from different sources and modalities in a unified manner.
For example, it can answer questions over multimodal data~\citep{liu2024visual,li2023blip},
use tools~\citep{yang2024can,schick2024toolformer,yao2022react},
use desktop apps and web~\citep{kapoor2024omniact,he2024webvoyager},
answer questions over long documents~\citep{zhao2024retrieval}.
These efforts have contributed to fundamentally transforming LLMs from merely a chatbot into a general-purpose agent that can interact with the real world and 
helps users in different ways.

Recently, these advances give rise to the idea\footnote{\href{https://www.youtube.com/watch?v=zjkBMFhNj\_g}{link}} of making LLM an operating system (OS), where one uses the LLM as a CPU of the OS that digests data from RAM which is the context window, and calls functions which are the external tools.
However, to accomplish such goal, one needs to address several key challenges.
\textbf{One of the most important features of an OS is that it is forever stateful}.
If permitted, it could store all data, software, and run logs, and can retrieve them when completing future tasks.
In comparison, so far, most of our interaction with LLMs is session-based and the LLMs are largely stateless across different sessions.

\begin{figure}[t]
    \centering
        \includegraphics[trim={0 4.8cm 8.5cm 0},clip,width=0.95\textwidth]{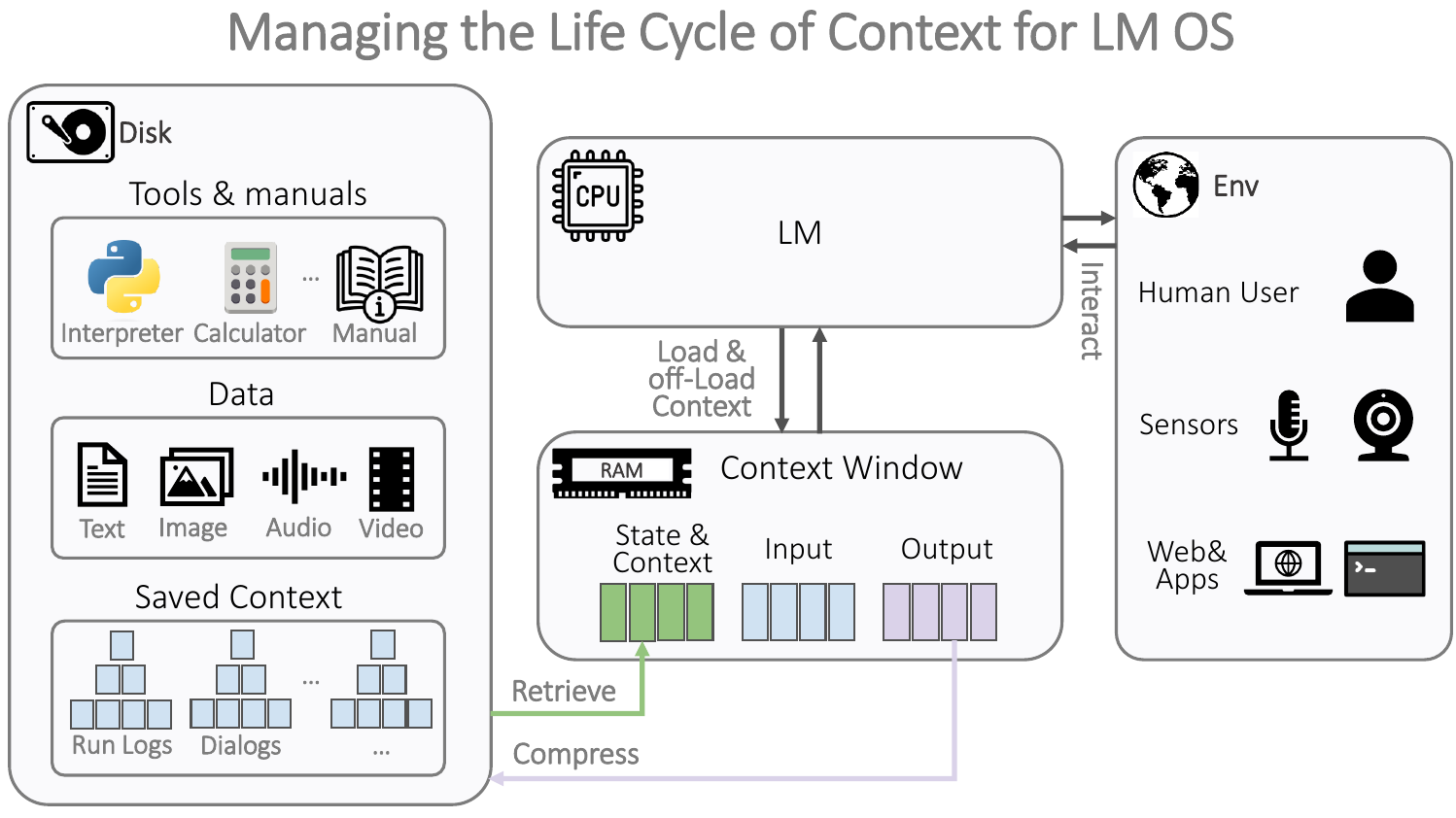}        
        \captionof{figure}{Building an LM OS requires a principled architecture to manage the life cycle of context.}
        \label{fig:lm-os-arch}
\end{figure}

This \textit{session-based paradigm} of interaction with LLMs results from several factors and one of them is the context window limit.
Recent LLMs are typically pre-trained with around 4K window size~\citep{dubey2024llama} (or higher for some proprietary models), with potentially long context fine-tuning up to 32K or more.
In inference time, some can scale to millions of input tokens such as Gemini\footnote{https://gemini.google.com/app}.
As large as it seems, it is still far from enough to digest context that could be fed to an OS for it to be meaningfully stateful:
A single HD image can take more than 1K tokens to represent;
one web search could return 10 web pages each with a few thousand tokens;
and a repo-level code can easily go up to thousands of lines.
While the session-based paradigm can already solve many problems, many daily tasks still require accessing long context.
To develop an LM OS that can assist with real-world tasks and stay stateful throughout the process, one must manage the context information in a life-long manner (Figure~\ref{fig:lm-os-arch}).
\textbf{We believe the lack of a principled architecture for managing the life cycle of context is one of the main obstacles to shifting from the session-based paradigm to the OS paradigm}.

In this preliminary work, we propose a novel architecture for managing the life-long context, namely the compressor-retriever architecture.
We design this architecture so that it stays model-agnositic and imposes minimal changes to the base model structure.
And, different from existing solutions such as retrieval-augmented generation (RAG), this architecture does not introduce standalone modules and relies only on the base model's forward function to compress and retrieve context, making the whole process end-to-end differentiable.
Specifically, the compressor builds a \textit{hierachical database} to store the chunked past context, where each chunk is represented by a coarse-to-fine memory hierarchy;
the retriever searches for relevant context with \textit{top-down sparse retrieval}, where it dynamically gathers context of different granularities with pure self-attention mechanism.
In our preliminary experiments, we validate this design in an in-context learning (ICL) reasoning task, where our model shows promising performance compared to the ideal setting.

\section{Related Work}
\label{sec:related-work}

While we draw our motivations from the LM OS, our work is also related to the long-context LLMs, where many efforts have been made to address the long-context challenge.
We categorize these works into two categories:

\textbf{Context compression}.
Context compression seeks to compress the context to reduce its size.
Some works focus on compressing the text explicitly into a condensed text; this includes Prompt-SAW~\citep{ali_prompt-saw_2024}, LLMLingua~\citep{pan_llmlingua-2_2024}.
Other works involve compressing the context in latent space, and this typically involves introducing a recurrent processing scheme, where the LLM processes the segments by reusing the compressed embedding from the last forward. This includes Transformer-XL~\citep{dai_transformer-xl_2019}, AutoCompressor~\citep{chevalier_adapting_2023}, ICAE~\citep{ge_context_2023}, and recurrent memory transformer~\citep{bulatov2022recurrent}.
Our work is closely related to the latter, from which we draw the inspiration for our compressor module.
However, these works are designed to digest the long context of a single session, and they lack a principled way to manage the context and agent states across sessions in a life-long manner.

\textbf{Retrieval-augmented generation}.
RAG is a widely used alternative for long context inference: rather than altering the base LLMs, it introduces a small standalone model (that is the \textit{indexer}) that splits the long context, e.g., a set of documents, into chunks and generates a vector index for them. During inference, the indexer retrieves the context by matching the current context with those chunks in the database.
RAG provides a straightforward way to manage the context, where one can directly add, delete, or change documents.
However, the performance is bounded by the capacity of the indexer model which is typically small and cannot be optimized together with the base model.
More importantly, it does not provide a state-dependent way to compress and retrieve context in different granularities.

\section{Method}

We propose a novel architecture that manages the life cycle of context in a principled manner.
A desired design should have the following properties:
(1) Handling infinite-length context;
(2) State- and task-dependent compression retrieval;
(3) End-to-end trainable without additional standalone modules.
The compressor-retriever architecture achieves the above by introducing two modules: compressor and retriever. Instead of introducing standalone adapters, we design modules to make extensive use of the base model's forward as the building blocks.
The idea behind is that we believe the base models, through massive pertaining, have already acquired the capability of compressing and reconstructing information, which can be elicited with supervised fine-tuning on the base model.
The other benefit is that since we only use the forward function, our architecture can be readily applied to all decoder-only transformer models.






\begin{figure}[t]
    \centering
        \includegraphics[trim={0 6.7cm 6.5cm 0},clip,width=\textwidth]{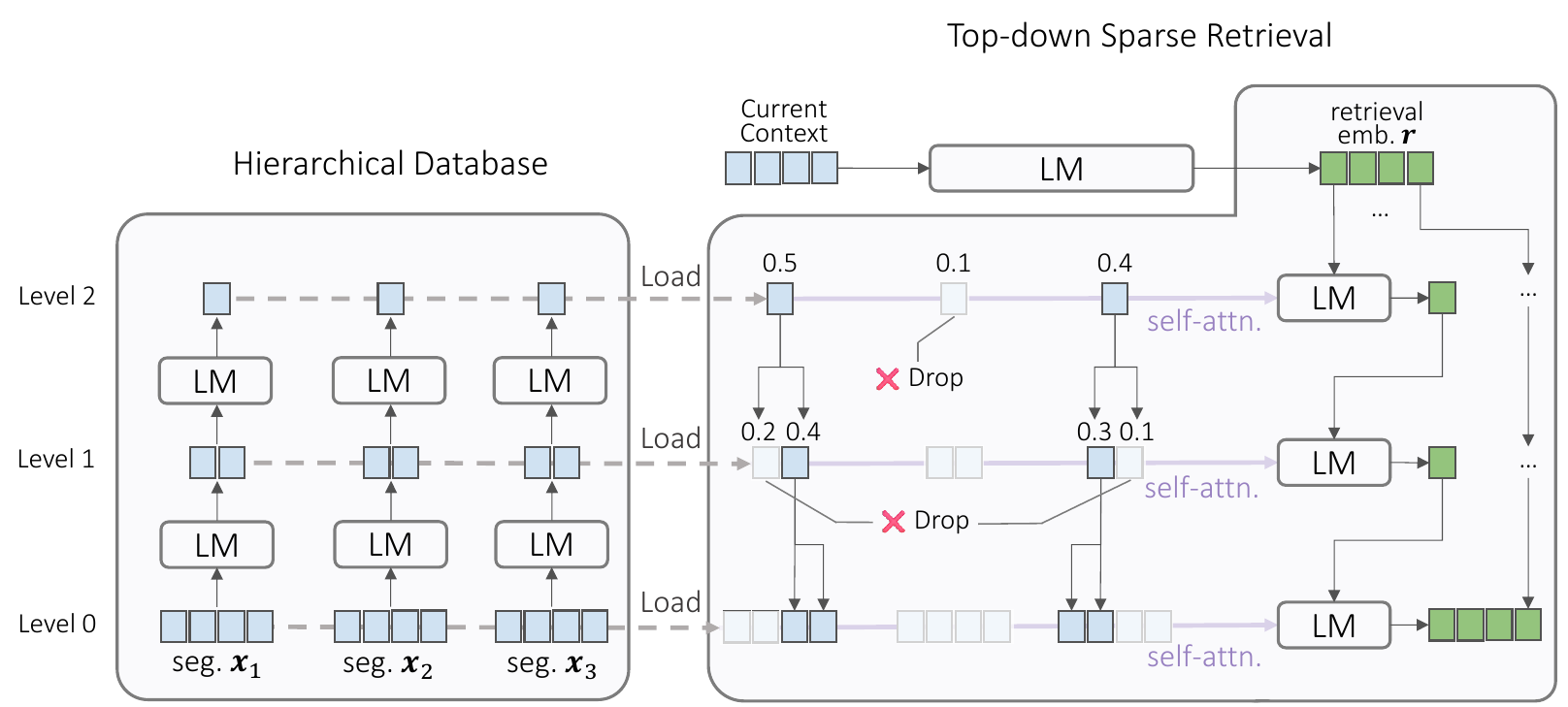}        
        \captionof{figure}{Overview of compressor-retriever architecture.}
        \label{fig:arch}
\end{figure}

\subsection{Compressor}

Let $\llm$ be the base LLM model.
The forward function $\llm(\vx)$ maps an input embedding sequence $\vx \in \bR^{n \times d}$ ($n$ tokens, latent dimension of $d$) to the output embedding of the same size $\vh \in \bR^{n \times d}$, where $\vh$ is the last hidden states.

The compressor module compresses the given context into a set of latent embeddings by appending the special $m = ``\mem\text{''}$ token to the context and obtaining the corresponding hidden states after a forward pass:
\begin{equation*}
    [\ \_\ ,\tilde{\vm}] \gets \llm([\vx, \vm], M). 
\end{equation*}
Let $k$ be the compression factor, 
$\vm= [m_1, m_2, ...]$ is the sequence of \mem\ tokens of size $n/k$, 
$\tilde{\vm} \in \bR^{(n/k) \times d}$ is the corresponding output hidden embeddings that encode the compressed information of $\vx$,
and $\_$ corresponds to the output of the sequence $\vx$, which we ignore.
$M$ is the special attention mask that we will introduce below.

\textbf{Retrieval-oriented compression}.
So far, the compressor resembles those compression techniques proposed in prior work such as AutoCompressor~\citep{chevalier_adapting_2023} and ICAE~\citep{ge_context_2023}.
However, we introduce two important features to enable a more flexible and life-long management of context information.

Instead of compressing and reusing the context only within a single session, we seek to build a \textbf{hierarchical database} that stores all external and past context, such as pages retrieved by web searches, the entire past chat history, logs generated by past tool usages, and so on.
Such a dataset should be built in a way where context can be retrieved efficiently and at different levels of granularity.

To this end, we build the hierarchical database by iteratively compressing the context to form an embedding hierarchy that encodes coarse-to-fine information.
Formally, given a segment $\vx$, we have:
\begin{align*}
    & \tilde{\vm}_0 \gets \vx, \\
    & [\ \_\ , \tilde{\vm}_{i+1}] \gets \llm([\tilde{\vm}_i, \vm_{i+1}], M_i),
    \ \ i=0, 1, ...
\end{align*}
For a fixed compression factor $k$, this forms a hierarchy
$[\tilde{\vm}_0, \tilde{\vm}_1, ..., \tilde{\vm}_L]$, where $L = \lceil \log_k n \rceil$.
Following this, one can build the database by processing the entire past context in the form of chunks $\vx_0, \vx_1, ...$, and the produced hierarchies can be stored on disk for future retrieval use.

\begin{wrapfigure}{r}{0.3\textwidth}
\centering
\includegraphics[trim={0 9cm 22.5cm 0},clip,width=\linewidth]{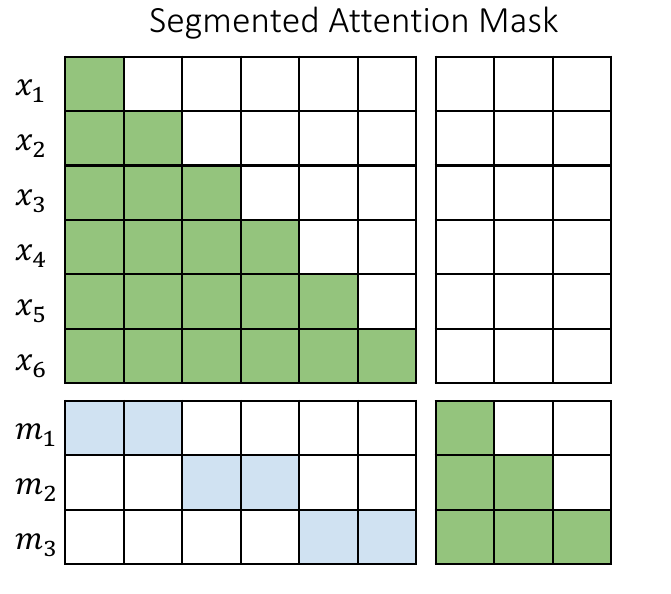}
\captionof{figure}{Segmented attention mask.}
\label{fig:segmented-attention-mask}
\end{wrapfigure}

\textbf{Segmented attention mask}.
During the retrieval phase, the model will conduct a top-down retrieval search to find the relevant context (details introduced later).
Therefore, it is necessary to compress the information in a \textit{structured} manner, so that the higher-level embeddings point to different parts of the lower-level embeddings.
To do so, we modify the standard causal attention mask to \textit{segmented attention mask}.
Specifically, let $[\vx, \vm]$ be the sequence to the compressed, and $m_j$ be the $j$th \mem\ token of $\vm$.
The standard attention mask lets $m_j$ attend to the full sequence $\vx$. However, if done this way, once $m_j$ is picked in the retrieval phase, it is unclear which part it attends to, making it difficult to narrow down the parts to be retrieved.
Instead, we make $m_j$ attend to a certain segment of $\vx$. There could be many ways to arrange the segmentation, and in this work, we first investigate a simple scheme, where $\vx$ is split sequentially into segments of length $k$, and $m_j$ attends to the segment $[x_{k*(j-1)}, ..., x_{k*j}]$.
We find this scheme is effective and we leave investigation on more advanced schemes in future work.
We illustrate the corresponding attention mask in Figure~\ref{fig:segmented-attention-mask}.

\subsection{Retriever}

We now show how to retrieve the context from the hierarchical database.
Similar to the compressor, we exclusively use the base model and its forward function in this process.
This sets us apart from prior work such as retrieval-augmented generation (RAG), which relies on an external small model to index and retrieve the context.
We argue that such a design can fully exploit the capability of the base model and impose least changes to the model's architecture, making it easy for fine-tuning.

The retrieval process starts by encoding the current context into retrieval embeddings which will be used to hold the retrieved information.
Let $\vx$ be the current context and $r = ``\ret\text{''}$ be the special retrieval tokens, we have
\begin{equation*}
    [\ \_\ , \tilde{\vr}] \gets \llm([\vx, \vr]),
\end{equation*}
where
$\vr = [r_1, r_2, ...]$ is the appended sequence of \ret\ tokens,
$\tilde{\vr}$ is the corresponding retrieval embeddings,
and $\_$ is the output corresponds to the $\vx$ sequence, which we ignore.
Unlike the compression case, the size of $\tilde{\vr}$ should be as large as possible so that more context could be retrieved, however, it should also be bounded by the context window size and leave space for next token generation.
For example, for a 4K window with 1K context, one could designate 1K for the retrieved context and the rest 2K for generation.
That said, retrieval embedding size is a hyperparameter that is largely dependent on the hardware resource and the nature of the task, which should be determined in a case-by-case manner.

\textbf{Top-down sparse retrieval}.
Now we retrieve the relevant context starting from $\tilde{\vr}$.
Recall that the database stores all the past context chunks $\vx_0, \vx_1, ...$, with each chunk also represented by a memory hierarchy $[\tilde{\vm}_0, \tilde{\vm}_1, ..., \tilde{\vm}_L]$.
Let $\tilde{r}$ be one of the retrieval embeddings, our core retrieval mechanism is to use $\llm$ to compute the attention scores of these memories with respect to $\tilde{r}$, and then aggregate them into $\tilde{r}$ level by level, from top to bottom, that is
\begin{equation*}
    \va_l, [\ \_\ , \tilde{r}_{l-1}] \gets \llm([\tilde{\vm}_l, \tilde{r}_l]),\ \ l = L, L-1, ..., 0,
\end{equation*}
where $\va_l$ is the last layer attention vector of $\tilde{\vm}_l$ with respect to $\tilde{r}_l$,
and $\_$ is the output corresponds to $\tilde{\vm}_l$, which we ignore.
Note that (1) this essentially performs a self-attention with memories and $\tilde{\vr}$;
(2) this means rather than retrieving the ``exact'' context (as that in RAG), we allow the model to re-compute the memory during retrieval, making it more flexible;
(3) following this implication and recalling that higher-level embeddings are effectively the higher-level (or more compressed) memories of the original chunk, one could early stop the search if the context is only remotely relevant or the desired granularity has been met.

With these observations in mind and that we know where the higher-level embedding attends to at its lower-level embeddings (thanks to the segmented attention mask), we further propose the \textit{sparse retrieval}:
\begin{align}
    & \tilde{\vm}_{l, C} \gets \text{TopC}(\tilde{\vm}_{l}, \tilde{\vm}_{l+1}, \va_{l+1}) 
    \label{eq:topc} \\
    & \va_l, [\ \_\ , \tilde{r}_{l-1}] \gets \llm([\tilde{\vm}_{l,C}, \tilde{r}_l])
    \label{eq:ret-self-attn} \\
    & l = L-1,\ ...\ , 0 . \nonumber
\end{align}
$\va_{l+1}$ is the attention of $\tilde{r}_{l+1}$ on the embedding at level $l+1$. We define the $\text{TopC}(\cdot)$ function, which will
(1) gather the top $C$ indices at level $l+1$ that has the highest attention score;
and (2) further gather the embeddings at level $l$ that were attended by the embeddings indexed by the top $C$ indices, making it $\tilde{\vm}_{l, C}$.
In~\eq{eq:ret-self-attn}, the $\tilde{\vm}_{l, C}$ of length $C*k$ (recall that each embedding attends to $k$ lower-level embeddings) becomes the input of lower-level self-attention together with the lower-level retrieval embedding $\tilde{r}_l$.

Intuitively, through~\eq{eq:topc} and~\eq{eq:ret-self-attn}, \textbf{we achieve a dynamic retrieval scheme that searches and aggregates context at different levels of granularities suiting the need of the current task with pure self-attention operations}.
At each level, we aggregate all context into $\tilde{r}_{l}$ through self-attention, making sure all information at this level of granularity is gathered.
And then based on the attention $\va_{l}$, we further identify the ``parts'' that the models are more interested in, which presumably are the context that are more relevant, and thus require more fine-grained information.
We collect the top $C$ of those and their lower-level embeddings and proceed with a lower-level search until we hit the bottom.


\section{Training, Inference, and Performance}

\subsection{Training}

The compressor-retriever architecture does not introduce additional standalone modules. While we believe the pre-trained LLMs have already acquired the necessary capabilities, it still requires fine-tuning to elicit such capabilities.
Setting up the training data and pipeline is nontrivial.
We will discuss the challenges and our proposed solutions in four aspects: parameters, training objectives, data, and performance considerations.

\textbf{Trainable parameters}.
The minimum parameters to train to run the architecture are the two embeddings for \mem\ and \ret\ tokens.
However, in experiments, we find the best performance is achieved by also fine-tuning the base LLM's parameters. This observation is also confirmed in related work~\citep{chevalier_adapting_2023, ge_context_2023}.
In our preliminary experiments, we find LoRA fine-tuning to be sufficient. However, one may as well do full parameter fine-tuning for optimal performance given enough hardware resources.
In summary, the trainable parameters are the LoRA adapters over all linear layers and the two embeddings.

\textbf{Training objectives}.
As the architecture is end-to-end differentiable, training the model to learn to compress and retrieve context becomes straightforward.
We use the standard autoregressive objective
\begin{equation*}
    \cL = - \frac{1}{n} \sum_{t=1}^n \log p(x_{t+1} | x_t, ..., x_1, \mM),
\end{equation*}
where the model predicts the next token given the previous context and the hierarchical database $\mM = [\tilde{\vm}_0, \tilde{\vm}_1, ...]$.
Intuitively, to predict the next token, the model learns to retrieve the most relevant and predictive context from the database.
In experiments, we find this objective is sufficient, which aligns with the observations in prior work~\citep{chevalier_adapting_2023}.
We leave investigations of other objectives such as the reconstruction loss~\citealp{ge_context_2023} in future development.

\textbf{Data}.
Training the model to effectively use the compression and retrieval capability, requires careful curation of the training and inference dataset.
Unlike the standard pertaining where sequences are typically 4K long, one must collect high-quality long-context data. However, compared to those short-context pertaining data such as The Pile~\citep{pile} and FineWeb~\citep{penedo2024finewebdatasetsdecantingweb}, native long-context data are almost non-existent.
To this end, many works resort to either privately collected data or synthetic data~\citep{dubey2024llama, deepseek-ai_deepseek-v2_2024}.
In our experiments, we sidestep this challenge by validating our architecture on a small-scale problem, and we leave this part in future development.

Apart from the data enabling the capabilities, we also need to collect data to teach the model to manage the context during inference. This is similar to a \textit{instruction fine-tuning} dataset.
For example, during inference, the model needs to know when to initiate the retrieval.
We plan to address this by constructing fine-tune dataset with special retrieval tokens \ttt{<call\_retrival>} inserted, where the model learns to decode such a token to initiate the process. Such an approach is also used in the dataset construction of prior tool-using works~\citep{schick2024toolformer,hao2024toolkengpt}.

\textbf{Performance considerations}.
Compared to the standard autoregressive training, our training is more expensive.
For a target sequence $\vx$, standard training performs forward only once with $\vx$ as the input to get the predicted labels, and only intermediate activations of this forward will be saved for backward pass.
Our case is to some extent similar to recurrent models such as RNN, as the label prediction is dependent on the output of the previous forward.
Specifically, let $\vc$ be the context of length $n$ and $k$ be the compression factor.
First, it takes $L = \lceil \log_k n \rceil$ times of forwards to build the hierarchy.
Then, for retrieval, it takes another $L + 1$ times to generate the retrieval embedding and search to the bottom.
Finally, a last call of forward is made to predict the labels of $\vx$.
In total, one needs to perform $2L + 2$ times of forwards before calling the backward.
Apart from the computation cost, this also leads to two other issues:

\textbf{1. Intermediate activations}:
PyTorch saves activations such as attentions and embeddings for the backward pass. Let $T$ be the total number of forwards called before backward, the space complexity is
$O(Tn^2 + Tnd)$.
In general, this cost is moderate compared to those long context training that goes up to 32K (e.g.,~\citep{dubey2024llama}), because it grows quadratically with the length of the context, and with our architecture, we can always limit the max length of every forward call to a small number (say, 4K or even smaller). Still, given that the total number of context chunks can be enormous (say, 10M tokens split into 5K chunks of length 2K), 
it is impossible to store all activations in VRAM.
In our experiments, we use gradient checkpointing to alleviate this issue. We also investigate implementing custom \ttt{pack} and \ttt{unpack} functions to dump chunk activations to RAM and disk, and only load them when they are in the top $C$ indices.

\textbf{2. Potential gradient instability}:
One of the main drawbacks of recurrent models is the gradient vanishing/explosion problem, as one needs to unfold the recurrent forward calls to compute the gradient, which is also referred to as \textit{backpropagation through time} (BPTT).
A widely used solution to this is the truncated BPTT, where one stops the gradient after certain times of unfolding. Such a technique is also used in prior work for transformer models~\citep{dai_transformer-xl_2019, chevalier_adapting_2023}.

Unfortunately, it cannot be applied in our case. Unlike seq2seq training used in these works, where every recurrent call has two sources of gradients (one from the prediction loss of the current state, and the other from BPTT from later states), the compression and retrieval calls happen in latent space and thus have no groundtruth labels, so the only source of the gradient is from the final autoregressive loss. This means one has to perform the complete BPTT to get the loss signal for compression and retrieval.
As bad as it sounds, so far we haven't encountered any gradient issues in our experiments. This is partially due to the depth $L$ grows in log scale with respect to $n$---For example, a 4K chunk with factor 4 leads to only 6 levels. So the total number of recurrent steps in our case is typically small.
Nevertheless, we also investigate using intermediate losses (e.g., reconstruction loss from compression embeddings) to train the compression and retrieval process to avoid performing the complete BPTT.

\subsection{Inference}

Inference with Compressor-Retriever architecture is straightforward.
A typical session begins with empty context or existing context similar to that in RAG use cases.
Given a context window size, we actively manage the window throughout the interaction.
For example, when the current context is about to exceed the window size, it automatically compresses the context and adds it to the database.
During the interaction with the user, the model can either be set to initiate retrieval whenever it is its turn to generate, or, after instruction fine-tuning, set to actively generate the \ttt{<call\_retrival>} token whenever it thinks appropriate. 

\textbf{Asynchronous retrieval}.
In either case, the retrieval process can hang the generation process, leading to a certain latency. Different from the training phase, where compression and retrieval happen before and synchronously with the generation, one can set both processes to run in the background and asynchronously with the generation.
For example, when the model decodes \ttt{<call\_retrival>} token, it initiates the retrieval in a separate process and continues its current generation, and the results will be gradually added to the current context as the generation goes.

\section{Experiments}

In this section, we validate the architecture in a small-scale preliminary experiment.

\textbf{Task}.
We evaluate our architecture on reasoning problems in an in-context learning (ICL) setting. Given a question $q$ the model is tasked to predict the answer $a$. Together with the target question, several examples $\anB{q, a}$ are included as the few-shot examples.
In our context, we consider all ICL examples $\vc = \{ \anB{q, a}_1, \anB{q, a}_2, ...\}$ are the context, and the target question $\vx = q$ is the input sequence.

If the model is never trained on the target dataset, having included ICL examples $\vc$ will lead to a significant increase in performance compared to the zero-shot case.
Therefore, we can validate our architecture by testing if the model could effectively compress and retrieve the ``right'' ICL examples for solving the current task.
Specifically, we provide the model with ICL examples from both the same dataset as the target question and datasets of other reasoning tasks. In other words, some ICL examples are relevant, while others are irrelevant.
We then limit the window size so that the model can only pick a subset of the examples.
If the model can successfully retrieve the relevant ones, then its performance will be similar to that of the full-example case.

\textbf{Data}.
We use four reasoning datasets to construct our training and testing datasets:
GSM8K~\citep{cobbe2021training} consists of math problems, FOLIO~\citep{han2022folio} consists of natural language inference problems, proScript~\citep{sakaguchi2021proscript} consists of graph reasoning problems, and ReClor~\citep{yu2020reclor} consists of commonsense reasoning problems.
These four tasks are sufficiently distinct~\citep{yang2024can}, so their ICL examples are irrelevant and contribute little to solving others' problems.
We construct ICL examples by concatenating their ground-truth answers to the questions.

\begin{wraptable}{r}{0.35\textwidth}
    \centering
    \begin{tabular}{@{}lc@{}}
    \toprule
    Mode                                                            & Accuracy \\ \midrule
    0-shot                                                          & 0.250    \\
    6-shot                                                          & 0.578    \\ \midrule
    \begin{tabular}[c]{@{}l@{}}Compressor\\ -Retriever\end{tabular} & 0.429    \\ \bottomrule
    \end{tabular}
    \caption{Accuracy of ICL reasoning tasks with base and compressor-retriever architecture.}
    \label{tab:demo-res}
\end{wraptable}

\textbf{Setting}.
We use \ttt{LLaMA3.1-8B-instruct} as the base model for the experiments. We choose the instruct version so that the model has a reasonable zero-shot capability, which could serve as a ``lower bound'' of the accuracy---If the model retrieves the right examples, its accuracy should be in between the zero-shot and the full-example cases.

Each sample in our dataset consists of 6 ICL examples: 2 from the same dataset as the target question, and 4 randomly sampled from the rest of the datasets.
For the training set, we have target sampled from one of GSM8K, proScript, and ReClor, and irrelevant samples from the other two;
for the test set, we have target sampled from only FOLIO, and irrelevant samples from one of GSM8K, proScript, and ReClor.
\textbf{Note that we only have FOLIO target problems in the test set and have deliberately excluded FOLIO examples from the training set}.
We believe that such compression and retrieval capabilities should be domain-agnostic, therefore, we restrict the training set to only contain ``out-of-distribution'' data (from the perspective of solving FOLIO problems), so that the model does not get trained on solving the FOLIO problem directly.
The resulting training set contains 3375 training samples with an average token length of 600;
the test set contains 192 samples with an average length of 514.
Due to limited GPU resources, we use a fixed compression scheme where the context is compressed to level 1 with 50 embeddings and then level 2 with 1 embedding.
All experiments are done on a 4090 GPU with 24GB RAM.
We fine-tune \ttt{LLaMA3.1-8B-instruct} on our training set with LoRA $r=8$ on all linear layers. We train it for 3 epochs with $lr=3e-4$, batch size 32 with a mini-batch size of 1.

\textbf{Results}.
We show results in Table~\ref{tab:demo-res}.
The experiment shows that our method achieves 75\% of the 6-shot ICL performance, indicating that the model has successfully picked the correct examples.
To further validate this aspect, we track the top-level attentions and compare the top indices with those of the relevant examples, the match rate is 64\%, meaning for 64\% of the test cases the model finds all the correct examples.
Still, the score can be significantly improved by scaling the experiments.
We leave this part in future development.



\section{Conclusion}

In this preliminary work, we propose the compressor-retriever, an architecture that manages the life cycle of context in a principled manner.
This architecture does not introduce standalone modules and can be trained end-to-end similar to the standard fine-tuning process.
Experiments show promising potentials of this architecture, which lays the foundation for developing the LM OS.

\bibliography{ref}
\bibliographystyle{plainnat}




\end{document}